\documentclass[preprint,12pt]{elsarticle}

\usepackage{amssymb}
\journal{the arXiv.org}

\begin{document}

\begin{frontmatter}

\title{Inference and Plausible Reasoning  in a Natural Language Understanding System Based on Object--Oriented Semantics}

\author{Yuriy Ostapov}
\address{Institute of Cybernetics of NAS of Ukraine, 40 Acad. Glushkova  avenue,  Kiev, Ukraine.
E-mail: yugo.ost@gmail.com }

\begin{abstract}

Algorithms of  inference in a computer system oriented to input and semantic processing of text information are presented.
Such inference is necessary for logical questions when the direct comparison of objects from a question and database can not give a result.
The following classes of problems are considered: a check of hypotheses for persons and non-typical actions, the determination of persons and circumstances for  non-typical actions,
planning actions, the determination of event cause and state of persons. 
To form an answer both deduction and plausible reasoning are used.
As a knowledge domain under consideration is social behavior of persons, plausible reasoning is based on laws  of social psychology.
Proposed algorithms of inference and  plausible reasoning can be realized in  computer systems closely connected with text processing (criminology, operation of business, medicine, document systems).
\end{abstract}

\end{frontmatter}

\section{Introduction}

The given paper is devoted to the solution of logical questions in a computer system based on a representation of natural language sentences using methods of mathematical logic (predicate calculus)\cite{Kleene}. 
{\it Persons, organizations,  machines, things,} and other objects are described with the help of predicates. Properties of these objects are presented with variables (arguments) of predicates. 
{\it Actions}  and {\it events} connected with these objects also are described with predicates, which contain references to given objects.

Such approach for natural language semantics was realized  in the system LEIBNIZ \cite{Ostapov1}.  
Algorithms of question answering in this system are considered in \cite{Ostapov2}.
An inference is inserted into these algorithms when the direct comparison of predicates from a question and database is insufficient  to  form an answer.

To get  an answer  the inference applies both deduction and plausible reasoning \cite{Polya}.
The use of plausible reasoning has been made possible when we  formulate  such reasoning  by means of logical rules.
We consider only some plausible reasoning based on laws of social psychology\cite{Mayers}.

Considering social behavior of persons (as a knowledge domain), first one should take account of {\it  role behavior} founded on common rules\cite{Mayers}.
These rules were formed on the base of collective experience and team-work.
Role behavior includes the execution of professional tasks, family relations, learning, medical treatment,  and the like.
Sometimes role behavior has game nature \cite{Bern}. Furthermore, there are acts of persons outside role  behavior. Such acts, as a rule, have motives demanding non-typical reaction.
On this basis, we consider the next classes of problems:

\begin{itemize}

\item a check of hypotheses for persons and non-typical actions;
\item the determination of persons and circumstances for  non-typical actions;
\item planning actions;
\item the determination of event cause and state of persons.
                                                                                  
\end{itemize}

The solution of these problems is founded on a {\it database} with the description of facts and  {\it knowledge base} including the description of {\it concepts, operations, scripts, diseases}, and {\it plans} (schemes).

Question answering using  inference and  plausible reasoning has a certain degree of validity, which depends on:

\begin{itemize}

\item reliability of primary facts;
\item reliability of plausible reasoning\footnote{Probabilistic assessment of plausible reasoning is considered in \cite{Polya}}.
                                                                            
\end{itemize}

The use of inference in  natural language understanding systems was studied in \cite{Carbonel, Fum, Green, Reiter, Schank, Smith, Velard}.
The essential distinction of our paper is that:\\[10pt]
\hspace*{20pt} 1. A natural language sentence is interpreted using  a set of predicates. These predicates describe persons, organizations, things, machines, and other objects as well as actions and change of state for these objects.\\
\hspace*{20pt} 2. The proposed approach is not attached to a certain design way. {\it  Visual Studio, Delphi, JBuilder}, and other program systems can be used as a design tool.   \\
\hspace*{20pt} 3. Information can be saved by means of modern database management systems: {\it  Oracle, Informix, MS SQL Server, DB2 }\footnote{Predicates are presented  with tables of a relational database.}. This permits to process the great volume of data (facts, dictionaries, knowledge).\\
\hspace*{20pt} 4. Any well-developed natural language can be selected to describe a knowledge domain.\\  [10pt] 

By this means, we are dealing with a new computer technology. It should be pointed out that the realization of this technology with the help of modern design tools and database management systems permits  to connect  algorithms 
under consideration with real computer systems for criminology, operation of business, medicine, etc.

\section{Knowledge base}

A knowledge base involves frames (articles) with the description of {\it concepts} (nouns and verbs), {\it operations, diseases, scripts}, and {\it plans} (schemes of cities, constructions, and the like).
The description of  concepts and operations was examined in  \cite{Ostapov1}. Therefore, we consider only the description of diseases, scripts, and plans.

\subsection{Description of scripts }

A history of life will be  referred to as a {\it script} when this history begins with a certain event and describes subsequent actions and state of persons and other objects. 
The script (for example, a {\it distress of ship}) is formed with the next sentences:\\  [10pt] 
{\it frame is the script of distress\\
ship has distress\\
persons sit down in boats\\
persons can be many days in sea\\
persons may die from lack of water and food}\\

\subsection{Description of disease }

The description of a disease (for example, {\it influenza}) is formed in the following way:\\  [10pt] 
{\it frame is influenza\\
there is high temperature\\
there is cough\\
there is headache}\\ 

It is suggested that the description of  characteristic can be accompanied with an appropriate attribute.

\subsection{Description of plans }
 
To solve real tasks persons use as well schemes, drawings, plans. Consider the description of a city plan.
In the system LEIBNIZ such information is presented with the predicates: {\it street, crossing, block, construction, apartment}.

The predicate {\it street} has the structure:\\[10pt]
\hspace*{20pt} 1. Code of street.\\
\hspace*{20pt} 2. Name.\\[10pt]

The predicate {\it crossing} has the next variables\footnote{It is used a coordinate system for a given city. Blocks attached to the given crossing are numbered clockwise starting with the top left block.}:\\[10pt]
\hspace*{20pt} 1. Code of crossing.\\
\hspace*{20pt} 2. First coordinate of crossing center.\\ 
\hspace*{20pt} 3. Second coordinate of crossing center.\\ 
\hspace*{20pt} 4. Code of first block. \\ 
\hspace*{20pt} 5. Code of second block. \\ 
\hspace*{20pt} 6. Code of third block.\\
\hspace*{20pt} 7. Code of fourth block.\\[10pt]

The predicate {\it block} has the structure\footnote{Streets and crossings  that  limit  the given block are numbered clockwise starting with the down left crossing.}:\\[10pt]
\hspace*{20pt} 1. Code of block.\\
\hspace*{20pt} 2. Code of first  street limiting the block.\\ 
\hspace*{20pt} 3. Code of second  street limiting the block.\\ 
\hspace*{20pt} 4. Code of third street limiting the block.\\ 
\hspace*{20pt} 5. Code of fourth  street limiting the block.\\ 
\hspace*{20pt} 6. Code of first crossing.\\
\hspace*{20pt} 7. Code of second crossing.\\
\hspace*{20pt} 8. Code of third crossing.\\
\hspace*{20pt} 9. Code of fourth crossing.\\
\hspace*{20pt} 10. First coordinate of block center.\\
\hspace*{20pt} 11. Second coordinate of block center.\\[10pt]

The predicate {\it construction} (house, theatre, station, and other)  has the next variables:\\[10pt]
\hspace*{20pt} 1. Code of construction.\\
\hspace*{20pt} 2. Number (for house).\\ 
\hspace*{20pt} 3. Code of street.\\ 
\hspace*{20pt} 4. Name (for theatre, station, and other).\\ 
\hspace*{20pt} 5. First coordinate of construction center.\\ 
\hspace*{20pt} 6. Second coordinate of construction center.\\
\hspace*{20pt} 7. Code of block.\\[10pt]

The predicate {\it apartment}  has the structure:\\[10pt]
\hspace*{20pt} 1. Code of apartment.\\
\hspace*{20pt} 2. Code of house.\\ 
\hspace*{20pt} 3. Number of entrance.\\ 
\hspace*{20pt} 4. Floor.\\
\hspace*{20pt} 5. Number of  apartment.\\[10pt]

\section{Check of hypotheses for persons and actions}

\subsection{Check of hypothesis for stay of person }

Such hypothesis can be examined using information  about person behavior or analysis of person movement:\\[10pt]
\hspace*{20pt} 1. Let us  check a person stay  at a given place and certain time on the base of person behavior\footnote {This task is considered in \cite{Ostapov2}.}. 
 In the case of failure we go to step 2.\\
\hspace*{20pt} 2. Let us analyze facts about a person stay at other locations and determine possibility  to go to the given place from other locations.\\[10pt]

First let us consider an algorithm of path determination from a starting-point to an ending-point.
If the starting-point and ending-point lie at the same block, we move along the shortest route.
The next algorithm is used to determine a path  from a starting-point to an ending-point when these points do not belong to the same block:\\[10pt]
\hspace*{20pt} 1. If the starting-point is a crossing, we go to step  3, otherwise to step 2.\\
\hspace*{20pt} 2. Let us determine a crossing (in the starting block) closest to the ending-point, move  to this crossing  along the shortest route, and go to step 3. \\ 
\hspace*{20pt} 3. Let us select a block attached to the given crossing when the distance between the block center and  the ending-point is minimal. 
If the given crossing and the ending-point lie at the selected block, then we move  along the shortest route and the algorithm is completed. 
Otherwise we go to step 4.\\ 
\hspace*{20pt} 4. Let us find a crossing at the selected block closest to the ending-point and move to this crossing through streets of the given block 
and go to step 3.\\[10pt] 

Using this algorithm, we can find the path length from the starting-point to the ending-point and determine possibility  to go to the given place from other locations.
To do this requires knowledge  of moving time, which is calculated with the help of average velocity for a car or pedestrian. 
The  average velocity of car depends on the street and selected point of time.

\subsection{Check of hypothesis for non-typical actions of person}
 
Non-typical actions, as a rule, have a set of motives or can be a result of strong emotional shock.
For example, a murder has serious motives: robbing, revenge, etc.

To check motives of non-typical action it is necessary for each motive to find a cause  that produces such motive.
For example, for a murder:\\[10pt] 
the motive {\it robbing} has  the cause {\it  subject is criminal};\\
the motive {\it revenge} has  the cause {\it insult of subject}.\\

The description of a non-typical action contains:

\begin{itemize}

\item person;
\item object of effect;
\item way or tool;
\item circumstances(motives and causes);
\item time and place.                                                                                  
\end{itemize}

One should use all these factors in an algorithm of hypothesis check  for non-typical actions.
The proposed  algorithm is founded on a {\it  basic fact} --- pointing a given action for an unknown person (in a database).
The description of such action in the basic fact contains as well an object of effect, location and time.   
This basic fact permits to simplify the formulation of a question as a time or location can be taken from the basic fact. 

The examination of hypothesis for non-typical action of a certain person includes:\\[10pt]
\hspace*{20pt} 1. A check of  this person stay at a pointed place.\\
\hspace*{20pt} 2.  A check of  motives  for this person to execute the given action. \\
\hspace*{20pt} 2.  A check of  ways to realize the given action by this person. \\[10pt]
 
The check of  this person stay at a pointed place was considered above.

The check of motive is founded on the description of motives for the given action in a knowledge base.
Selecting the description of motive, we examine if the cause for this motive corresponds to a checked person.

The check of ways is based on the description of these ways for a given action in the knowledge base.
This check is realized  for a given person as follows:\\[10pt]
\hspace*{20pt} 1. If a way is contained in the question, then the use of this way in actions of the given person is examined by means of a database.
Also it is verified if this way corresponds  to the action from the question.\\
\hspace*{20pt} 2. If a way is not contained in the question, then we select actions of the checked person using the database and examine if  ways of these actions correspond to the way for  the action from the question.\\
\hspace*{20pt} 3. If the first and second items do not give the result, then  the description of way from the basic fact is used.
Any action of checked person are selected  from the database, and ways of these actions are compared with the way from the basic fact. 
If there is such way in an action from the database, then it is examined if this way corresponds to the action from the question.\\[10pt]

By way of illustration, let us consider the next example describing a murder. Let a database consist of the frames for a certain city and year. 
All further examples of questions correspond to this database.

Let us input  the next sentences into the database:\\[10pt] 
{\it 
a) The man shot a girl at 20 o'clock on the seven of November in 9 Street1 Street.\\
b) Petrov met a friend in 9 Street1 Street.   He bought a cheese after 19 o'clock.\\
c) Petrov is criminal.\\
d) Perov has a pistol.}\\

The sentence {\it a)} describes a basic fact, and the sentence {\it  b)} actions of  a person.
The sentence {\it  c)} notes that this person has a cause for the motive {\it  to rob}. This cause corresponds to the frame {\it   to shoot a person} from the knowledge base.
The sentence {\it  d)} points that the person has  the tool of murder. This tool corresponds to the description of tool in the same frame.

The system LEIBNIZ gives the positive answer for the questions:\\[10pt] 
{\it   Did Petrov shoot a girl on the seven of  November?\\
Did Petrov shoot a girl in 9 Street1 Street?}\\

\section{Determination of persons and circumstances for non-typical actions}

To determine a person and circumstances for non-typical actions the above-mentioned algorithms  are used  to check a person stay, motives and  ways of action. 
As noted above, proposed algorithms apply also a basic fact. 

\subsection{Determination of person for a non-typical action}

To determine a person  the next algorithm is used:\\[10pt]
\hspace*{20pt} 1. Persons are selected  from a database using a time and location of action from a question or basic fact.\\
\hspace*{20pt} 2. It  is examined for each person and the given action if a motive and tool  taken from a knowledge base correspond to this person.\\
\hspace*{20pt} 3. If  the given person satisfies all the checks, then an answer contains this person.\\[10pt]

The system LEIBNIZ gives a positive answer  for the question:\\[10pt] 
{\it   
Who shot  a girl in 9 Street1 Street?}\\

This answer will be {\it Petrov}.

\subsection{Determination of cause for non-typical action}

To determine a cause of a non-typical action the next algorithm is used:\\[10pt]
\hspace*{20pt} 1. A check of a given person stay is executed  using a time and location of action  from a question or basic fact.\\
\hspace*{20pt} 2. It is  examined for the given person and action if a motive and tool   taken from a knowledge base correspond to this person.\\
\hspace*{20pt} 3. If  the given person satisfies all the checks, then an answer contains a cause (for the motive) that is taken as well from the knowledge base.\\[10pt]

The system LEIBNIZ gives a positive answer for the question:\\[10pt] 
{\it 
 Why did Petrov  shoot   a girl  in 9 Street1 Street?}\\

This answer will be {\it as subject is criminal}.

\subsection{Determination of way for non-typical action}

To determine a way for a non-typical action the next algorithm is used:\\[10pt]
\hspace*{20pt} 1. A check of a given person stay is executed  using a time and location of action from a question or basic fact.\\
\hspace*{20pt} 2. It  is  examined for the given person and action if  a motive and tool taken from a knowledge base correspond to this person.\\
\hspace*{20pt} 3. If  the given person satisfies all the checks, then an answer contains a way of  action that is taken as well from the knowledge base.\\[10pt]

The system LEIBNIZ gives a positive answer   for the question:\\[10pt] 
{\it     
  How  did Petrov  shoot   a girl  in 9 Street1 Street?}\\

This answer will be {\it  by pistol}.

\section{Planning actions}

Consider the following problems:

\begin{itemize}

\item finding a person that, probably, plans a pointed operation;
\item the determination of a possible operation planed by a certain person;
\item the determination of ways and  actions  to execute a given operation by a certain person.
                                                                       
\end{itemize} 

The solution of these tasks will be illustrated by the example of the operation {\it robbing an office}. A knowledge base contains the description of this operation\footnote{We use only one alternative.}:\\[10pt] 
{\it   frame  is  rob office\\
alternative 1; to go to office\\
alternative 1; to come in through window  if  signalling does not work\\
alternative 1; to open safe with tool\\
alternative 1; to take money\\
alternative 1; to come out  through window }\\

\noindent as well as the description of the action {\it  to rob} and its motive:\\[10pt] 
{\it     frame  is  rob\\
 to take money as subject is criminal}\\

\subsection{Finding a person that plans a pointed operation}

First let us consider a check of an operation for a given person --- a subject of the operation.  For  a certain alternative (starting with the first), each stage is examined:
\begin{itemize}

\item if a condition describes  a certain situation that provides  the execution of  the stage, then we examine possibility to transmit this information to the subject of the operation;
\item if a condition concerns  the subject  of the operation, then  we examine  possibility to execute  a described action by the given person;
\item  if the stage contains a way, then we check if  the subject of the operation manages this way.
                                                                       
\end{itemize} 
 
If all stages of the checked alternative satisfy given conditions, then the check of the operation is completed. Otherwise the next alternative is examined.

To find  a person that, probably, plans a pointed operation:\\[10pt] 
\hspace*{20pt} 1. Let us select  persons  from a database if  these persons have the same motive as for  the planed action of the operation.\\
\hspace*{20pt} 2. All alternatives are examined for each selected person. If an alternative satisfies all conditions for a given person then a result is this person.\\[10pt]

Consider our example  {\it robbing an office}. An answer will be  {\it Petrov} for the question  {\it  Who plans (intends, wants) to rob the office in 9 Street1 Street?}. 

\subsection{Determination of a possible operation planed by a certain person}

A sequence of this problem solution is next:\\[10pt] 
\hspace*{20pt} 1. Let us select an operation from a knowledge base and examine all alternatives as it was described above.
 If a check of all alternatives does not give a result, then we go to the next operation.\\
\hspace*{20pt} 2. If the check of the operation  has  the positive result, then we examine motives of a  given person for the execution of  this operation.\\
\hspace*{20pt} 3. If the check of motives  gives the positive result, then the algorithm is completed, and a result is the name of the operation.
Otherwise we go to the check of the next operation.\\[10pt]

An answer can  be {\it to rob office}  for the questions:\\[10pt] 
{\it   What  does Petrov plan  in 9 Street1 Street?\\
 Which(what) operation  does Petrov plan  in 9 Street1 Street?}\\

\subsection{Determination of ways and  actions  to execute a given operation  by a certain person}

A sequence of this problem solution is next:\\[10pt] 
\hspace*{20pt} 1. Let us examine all alternatives for a given operation as it was described above.\\
\hspace*{20pt} 2. If the check of the operation has the positive result, then we examine motives of a given person for the execution of   this operation.\\
\hspace*{20pt} 3. If the check of motives  gives the positive result, then the algorithm is completed, and a result contains the description of actions and ways taken from a knowledge base.\\[10pt]

An answer can  be:\\[10pt] 
{\it  to come in through window\\
to open safe with tool}\\[10pt] 
 for the question {\it  How does Petrov plan (intend, want) to rob the office  in 9 Street1 Street?} \\[10pt] 

\section{Determination of event cause and person state }

To determine a cause of an event:\\[10pt] 
\hspace*{20pt} 1. Let us select events from a database  if these events have the same state, time, and place as in a question.\\
\hspace*{20pt} 2. Let us find an action (in the database) that produces the given event.\\[10pt] 
 
We will get the answer  {\it  as man shot girl}  for the question {\it  Why is a girl dead on the seven of November in 9 Street1 Street?.}

A state of a person depends on:

\begin{itemize}

\item   previous events that are accompanied with certain consequences;
\item   internal physiological and psychological factors\cite{Mayers}.
\
                                                                       
\end{itemize} 
An algorithm for the determination of  person state is next:\\[10pt] 
\hspace*{20pt} 1. If lately there was an action that  produces an event connected with the state of a given person\footnote{It is pointed in the {\it  dictionary of verbs.}}, then an answer is such state.\\
\hspace*{20pt} 2. If  there is a symptom of a disease\footnote{Symptoms  of the  disease are described in an appropriate frame of a knowledge base.}, then an answer is this disease.
\hspace*{20pt} 3. If a situation in which a person was after a late event is typical for a certain script\footnote{This script must be incorporated in a knowledge base.}, then an answer is the person state from this script.\\[10pt]

\section{Conclusion}

To solve the natural language understanding problem for a certain knowledge domain we apply the ontological approach. 
As a natural language sentence describes  facts (actions and events for objects of the real world), the problem of the given sentence understanding is in an adequate representation  of these facts.
Such representation is realized in our paper with the help of  predicate calculus.

Each posed task from the knowledge domain  also is interpreted using a set of predicates. Validity of an answer is examined by  an expert.
Possibility of such task solution is the problem of  solvability (for general questions) or computability (for special questions)\cite{Kleene}.

Typical tendency for modern computer systems is  in the use of artificial intelligence algorithms\cite{Russel}. 
Furthermore,  the formulation of problems using a natural language will be logical in the context of this tendency.
It should be emphasized that  {\it  the application of a natural language  for the description of problems essentially extends  the domain of solvable tasks at the cost of the most actual problem inclusion}.

Consider, for example, the problem of enterprise management.  At the moment  the technology ERP (Enterprise Resource Planning) is applied \cite{Chase}.
However, there is no way in such systems to enter immediately natural language questions  for diagnostics of financial state, prediction and planning of enterprise activity. 
To find  answers for pointed tasks it is necessary to execute many sequential  questions to a database and to process selected data manually.
Consequently,  the solution of the most important problems of enterprise management depends on qualification and honesty of available specialists.
Therefore, it is desirable to automate these tasks in addition to ERP. The experimental system of such type is described in  \cite{Enter} .

Other example concerns the domain of criminology.  Living computer systems save information about criminal offences by means of databases\cite{Manning} .
However, this information can not be used in full measure for logical processing as algorithms of semantic analysis are  not applied.
Thus, many actual problems are solved only by persons. Proposed algorithms permit to create a more intellectual  system based on  methods of criminalistics\cite{Belkin}.   
Such system will  form answers for natural language questions about criminal offences using the purposeful selection from a database and logical processing of selected data.
Some algorithms of such task solution are considered in this paper.

By this means, the presented technology discovers the perspective for the successful  solution of problems  in the social domains  with the help of intellectual computer systems.

\end{document}